\documentclass{bmvc2k}
\usepackage{booktabs}
\usepackage{bbm}
\usepackage{amssymb}
\usepackage{enumitem}
\usepackage[capitalize,nameinlink]{cleveref}
\usepackage{longtable}
\newcommand{\N}{\mathcal{N}}
\newcommand{\MB}{\mathcal{B}}

\newcommand{\lalign}{L_{a}}
\newcommand{\luniform}{L_{u}}

\usepackage{hyperref}

\hypersetup{
    colorlinks=true,
    linkcolor=red,
    filecolor=magenta,      
    urlcolor=cyan,
    pdftitle={Overleaf Example},
    pdfpagemode=FullScreen,
    }

\usepackage{bm,amsmath}
\usepackage{xspace}
\usepackage{graphicx}
\usepackage{color}

\usepackage{colortbl}
\definecolor{light_gray}{RGB}{170,170,170}

\usepackage[utf8]{inputenc} 
\usepackage[T1]{fontenc}    
\usepackage{url}            
\usepackage{booktabs}       
\usepackage{amsfonts}       
\usepackage{nicefrac}       
\usepackage{longtable,lscape}
\usepackage{array,multirow}
\usepackage{enumitem}

\usepackage{wrapfig}

\newcommand{\hide}[1]{}

\newcommand{\fb}{\mathbf{f}}

\newcommand{\hb}{\mathbf{h}}

\newcommand{\xb}{\mathbf{x}}
\newcommand{\yb}{\mathbf{y}}
\newcommand{\zb}{\mathbf{z}}

\title{Correlation between Alignment-Uniformity \\ and Performance of Dense Contrastive Representations}
\addauthor{Jong Hak Moon Student}{jhak.moon@kaist.ac.kr}{1}
\addauthor{Wonjae Kim Collaborator}{wonjae.kim@navercorp.com}{2}
\addauthor{Edward Choi Prof}{edwardchoi@kaist.ac.kr}{1}

\addinstitution{
 KAIST, Daejeon, South Korea
}
\addinstitution{
 Naver AI, Sungnam, South Korea
}

\runninghead{Student, Collaborator, Prof}{BMVC Author Guidelines}

\usepackage{soul}
\def\eg{\emph{e.g}\bmvaOneDot}

\begin{document}
\maketitle

\begin{abstract}
Recently, dense contrastive learning has shown superior performance on dense prediction tasks compared to instance-level contrastive learning. Despite its supremacy, the properties of dense contrastive representations have not yet been carefully studied. Therefore, we  analyze the theoretical ideas of dense contrastive learning using a standard CNN and straightforward feature matching scheme rather than propose a new complex method. Inspired by the analysis of the properties of instance-level contrastive representations through the lens of alignment and uniformity on the hypersphere, we employ and extend the same lens for the dense contrastive representations to analyze their underexplored properties. We discover the core principle in constructing a positive pair of dense features and empirically proved its validity. Also, we introduces a new scalar metric that summarizes the correlation between alignment-and-uniformity and downstream performance.
Using this metric, we study various facets of densely learned contrastive representations such as how the correlation changes over single- and multi-object datasets or linear evaluation and dense prediction tasks.
The source code is publicly available at: \text{https://github.com/SuperSupermoon/DenseCL-analysis}
\end{abstract}

\section{Introduction}
\vspace{-2mm}
 Instance-level CL (Contrastive Learning) with a single-object dataset (\eg ImageNet \citep{imgnet}) \citep{simclr, swav, misra2020self, moco, selfie, caron2018deep} has shown to be highly effective for learning visual representations in a self-supervised manner. To understand the semantic structures and behavior of this method, a few recent studies \citep{wang2020understanding,chen2021intriguing} analyzed the latent space (\eg unit hypersphere) from the perspective of uniformity and alignment (closeness).
 Intuitively, it is effective to analyze from these two perspectives, since features of all classes can be linearly separated from the rest of the feature space if they are sufficiently well clustered. 

\footnotesize
\begin{figure}
\centering
\begin{tabular}{cc}
\bmvaHangBox{{\includegraphics[width=5.5cm]{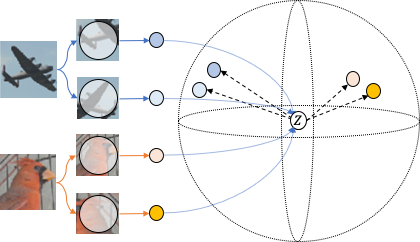}}}&
\bmvaHangBox{{\includegraphics[width=5.5cm]{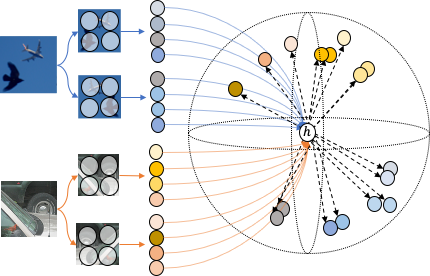}}}\\
(a) Instance-level CL & (b) Dense CL
\end{tabular}
\vspace{+2mm}
\caption{\textbf{Contrastive Representations on the hypersphere.} We demonstrate the difference in feature representation between instance- and dense CL on (a) single-object and (b) multi-object datasets. (a) represents an image as a single feature vector $\zb \in \mathbf{R}^d$ containing global information, whereas (b) represents a set of vectors $\hb \in \mathbf{R}^{d\times HW}$ exploited from a $H \times W$ feature map containing local feature information.}
\label{fig:teaser}
\vspace{-1mm}
\end{figure}
\normalsize


Although instance-level contrastive features have been successful in improving image classification performance, it has been observed that they do not enjoy the same transferability to dense prediction tasks (\eg object detection tasks) \citep{DenseCL, Propa, VADeR, SetSim, imgnet_rethink, insloc, swav, simclr}.
Since the receptive field of global averaged pooled features typically extends to the entire image, the pooled features are affected by background information, making it difficult to localize.
To overcome this gap, recent studies \citep{DenseCL, Propa, VADeR, SetSim} have developed dense CL with multi-object datasets (\eg MS-COCO \citep{coco}), using dense features to explicitly consider spatial information over regions and achieved comparable or better results compared to supervised ImageNet pre-training.
Despite such initial success, these works beg an important yet unexplored question: "How different are the dense-level features compared to the instance-level features?" (\hyperref[fig:teaser]{Fig.1})
In this work, we investigate the dense feature representation in terms of alignment and uniformity inspired by the pioneering analyses of \citep{wang2020understanding, chen2021intriguing}.
We extend the conventional contrastive loss (InfoNCE \citep{infoNCE}) to construct a more principled dense-level contrastive loss, and introduce a scalar metric to succinctly report the alignment-uniformity behavior of latent features.
Based on extensive experiments and analysis using both single and multi-object pre-training datasets, and instance-level (\textit{i.e.} linear evaluation) and dense downstream task (\textit{i.e.} object detection), our findings and contributions can be summarized as follows:
\begin{itemize}[leftmargin=4.5mm]
    \item We empirically show that the alignment-uniformity property in dense features is correlated with both instance-level and dense-level downstream task performance.
    \item We find that, contrary to our belief, instance-level contrastive features pre-trained on multi-object dataset can perform well on object detection, and dense contrastive features pre-trained on single-object dataset can perform well on linear evaluation, both cases following the alignment-uniformity principle.
    \item We discover the core principle in constructing a positive pair of dense features and empirically proved its validity with a simple index-wise matching.
\end{itemize}
\vspace{-2mm}

\section{Related work}
\vspace{-2mm}
After the advent of SimCLR \citep{simclr}, unsupervised CL (contrastive learning) was explosively researched on the instance-level \citep{swav, misra2020self, moco, selfie, caron2018deep}.
The core idea of this approach is sharing the InfoMax \citep{linsker1988self} principle under instantiation by maximizing mutual information between two transformed versions of the same image \citep{CMC, mutual_info1, mutual_info2}.
Recently \citet{wang2020understanding} empirically proved that a unit $l2$-norm constrained contrastive loss (InfoNCE \citep{infoNCE}) can be decomposed into a metric of alignment ($l2$-distance) and uniformity (average pairwise Gaussian potential).
Also, they proved that optimizing contrastive loss is equivalent to optimizing the alignment between positive pairs and maintaining uniformity across all feature vectors in the hypersphere, and observed optimizing the alignment-uniformity properties is closely related to the downstream task performance such as linear evaluation.
This hypersphere uniform distribution was generalized by \citet{chen2021intriguing} and extended to a wider set of prior distributions (\eg uniform hypercube or normal distribution).
Our study is more related to \citet{wang2020understanding}, and we extend this analysis to dense features that contain local spatial information.
Recently, \citet{imgnet_rethink, tan2020efficientdet, sun2019deep} demonstrate a transfer learning gap between instance-level pre-training and dense prediction tasks such as object detection.
In an effort to overcome this gap, several works \citep{DenseCL, Propa, VADeR, SetSim}  generalized the instance discrimination from image-level to pixel-level to explore dense-level unsupervised CL and demonstrated improved downstream performance for dense prediction tasks.
In contrast to numerous theoretical \citep{theo1, theo2, theo3, theo4, theo5} and empirical analyses \citep{emp1, emp2, emp3, emp4, emp5, wang2020understanding, chen2021intriguing} to understand instance-level CL, no attempt has been made to understand dense CL.
While there are many open questions, in this work we analyze how the pre-training impacts downstream tasks by extending the instance-level contrastive loss to the dense-level paradigm.
Additionally, unlike instance-level CL where positive pairs are easily constructed via augmentations, constructing positive dense feature pairs in dense CL is non-trivial.
Each of the previous works devised its own strategy to solve this problem, such as calculating the cosine similarity between dense features \citep{DenseCL}, attention-based set-wise matching \citep{SetSim}, and matching dense features with associated regions \citep{VADeR, Propa}.
In this work, we take a more straightforward approach and adopt an index-wise matching between dense features from two augmented views.
In the experiments section, we compare this rather simple strategy with more sophisticated ones such as using cosine similarity or optimal transport, and report that our approach leads to comparable or better downstream performance.
Furthermore, we analyze the effectiveness of the index-wise pairing strategy in terms of whether the pre-training dataset consists of single-object images or multi-object images.
\vspace{-2mm}

\section{Method}
\vspace{-2mm}
\subsection{Preliminary: Instance-level Contrastive Loss}
\label{ssec:instcl}
Instance-level CL can be seen as the lower bound of mutual information ($MI$) between a positive pair $x$ and $y$ \cite{infoNCE, mutual_info1, mutual_info2}.
Given $MI(x,y) = H(x) \mathbin{-} H(x|y)$, the two right-hand side terms can be linked to the following two properties \citep{wang2020understanding, chen2021intriguing}:
\vspace{-2mm}
\begin{itemize}[leftmargin=5.5mm]
    \itemsep 0em 
    \item[$\ast$] Uniformity $H(x)$: Maximizing entropy leads to uniformly distributed latent vectors.
    \item[$\ast$] Alignment $H(x|y)$: Minimizing conditional entropy given the positive pair of each item makes them be aligned in the latent space.
\end{itemize}
\vspace{-2mm}
Note that the general form of contrastive loss is defined as follows,
\footnotesize
\begin{equation}
    \label{eq:instcl}
    L^{InsCont} \text{=} \mathbin{-}\frac{1}{\mathcal{N}}\sum_{i,j \overset{\mathrm{i.i.d}}{\sim} \MB}\log\frac{e^{sim(\zb_i,\zb_j)/\lambda}}{\sum_{k \overset{\mathrm{i.i.d}}{\sim} 2\mathcal{N}}\mathbbm{1}_{[k \neq i]}e^{sim(\zb_i,\zb_k)/\lambda}} ,\quad sim(\xb, \yb) \text{=} \frac{\xb \cdot \yb}{\lVert \xb\rVert\lVert \yb\rVert}
\end{equation}
\normalsize
where $\mathcal{N}$ denotes the number of randomly drawn instances, $\MB$ the minibatch, $\zb_i$ and $\zb_j$ the positive pair of instance-level latent vectors projected into a hypersphere, $\lambda$ the temperature, and $\mathbbm{1}_{[k \neq i] \in {0,1}}$ an indicator function.
\hyperref[eq:instcl]{Eq. (1)} can be rewritten as follows by applying logarithmic rules:
\footnotesize

\begin{figure*}
\centering
\includegraphics[width=8cm]{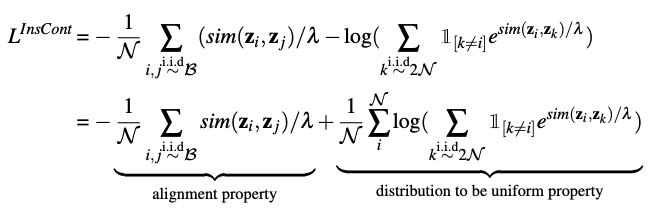}
\footnotesize
\end{figure*}
\normalsize
where we confirm that the contrastive loss indeed consists of two objectives.

\subsection{Dense Contrastive Loss}
\label{ssec:densecl}
In order to analyze the behavior of dense features in CL, we first formalize the dense CL objective, a natural extension of instance-level CL to the dense-level. 
Let $f$ be a CNN encoder that transforms an input image $x$ to dense feature vectors $\hb=f(x)=\{\hb_1, \hb_2, \ldots, \hb_{HW}\}, \enskip \hb_i \in \mathbb{R}^{d}$, where ${HW}$ is the spatial dimension size.

Following the principle of $MI$ maximization in \hyperref[eq:instcl]{Eq. (1)}, we assume that all $\hb_i$'s in a single image are \textbf{\emph{$i.i.d$}}. Although $\hb_i$'s do share some global information, this assumption is based on the fact that the values of each $\hb_i$ are not identical because each contains different spatial information. Also, this assumption is often implicitly seen in the previous dense CL studies to extract the corresponding feature. In particular, DenseCL\cite{DenseCL} compares all individual cosine similarity scores of features and pulls the most similar pairs closer. Also, Setsim\cite{SetSim} matches the corresponding feature set by calculating the set similarity using the attention score of the individual features. Therefore, by following the implicit \textbf{\emph{$i.i.d$}} assumption of the latest studies above, we perform index-wise feature matching by assuming \textbf{\emph{$i.i.d$}} of the output feature to form positive and negative pairs.

Dense contrastive loss can be defined as follows:
\footnotesize
\begin{equation}
    \label{eq:densecl}
    L^{DenseCont} \text{=} \mathbin{-}\frac{1}{\mathcal{N}}\sum_{i,j \overset{\mathrm{i.i.d}}{\sim} \MB}\frac{1}{HW}\sum_{p}^{HW}\log\frac{e^{sim(\hb_{(i,p)},\hb_{(j,p)})/\lambda}}{\sum_{k \overset{\mathrm{i.i.d}}{\sim} 2\mathcal{N}}\sum_{q}^{HW}\mathbbm{1}_{[k \neq i] \times [{\frac{q \neq p}{k \text{=} j}}]}e^{sim(\hb_{(i,p)},\hb_{(k,q)})/\lambda}} 
    ,\quad sim(\xb, \yb) = \frac{\xb \cdot \yb}{\lVert \xb\rVert\lVert \yb\rVert}
\end{equation}
\normalsize
where $\hb_{(i,p)}$ indicates $p$-th dense feature of the $i$-th sample, and $\mathbbm{1}_{[k \neq i] \times [{\frac{q \neq p}{k \text{=} j}}] \in {0,1}}$ an indicator function.
Note that a positive pair of dense features in our formulation consists of two dense features from the same index (\textit{i.e.} spatial position) of each augmented image pair (see the numerator of \hyperref[eq:densecl]{Eq. (2)}).
We discuss the strategy for choosing positive and negative dense pairs in further detail in \hyperref[ssec:dfm]{Section 3.3}.
\hyperref[eq:densecl]{Eq. (2)} can also be rewritten as follows by applying logarithmic rules:
\footnotesize

\vspace{-3mm}
\begin{figure*}[!ht]
\centering
\includegraphics[width=11.7cm]{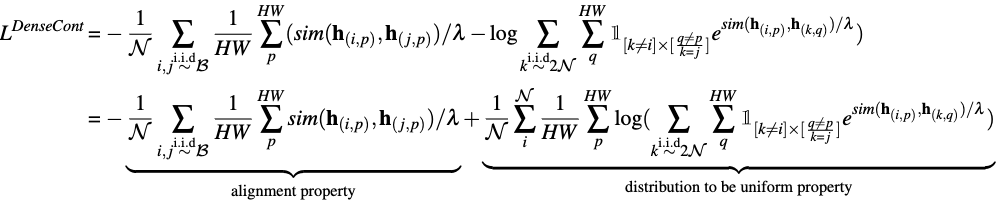}
\footnotesize
\end{figure*}
\vspace{-3mm}
\normalsize
where we again observe that dense CL consists of alignment and distribution objectives.
Therefore, by optimizing \hyperref[eq:densecl]{Eq. (2)}, dense features will asymptotically achieve the alignment-uniformity properties, similar to the instance-level CL.

To control these properties more directly, we adopt the metrics proposed in \citet{wang2020understanding} and extend them to the dense-level.
For the uniformity loss, we utilized a Gaussian potential kernel $G: \mathcal{S}^d \times \mathcal{S}^d \rightarrow \mathbb{R}_+$ \citep{wang2020understanding,borodachov2019discrete,cohn2007universally} and the logarithm of the dense average pairwise Gaussian potential.
Dense-level alignment-and-uniformity loss can be defined as:
\footnotesize
\begin{align*}
    \lalign \triangleq \mathbin{-}\frac{1}{\mathcal{N}}\sum_{i,j \overset{\mathrm{i.i.d}}{\sim} \MB}\frac{1}{HW}\sum_{p}^{HW}sim(\hb_{(i,p)},\hb_{(j,p)}),\quad
    \luniform \triangleq log \frac{1}{\mathcal{N}}\sum_{i,j \overset{\mathrm{i.i.d}}{\sim} \MB}\frac{1}{HW}\sum_{p}^{HW}G(\hb_{(i,p)},\hb_{(j,p)})
\end{align*}
\normalsize
where $G(\xb,\yb) = e^{-{\lVert \xb - \yb\rVert}^2_2}$.
denotes a pairwise Gaussian potential.

Perfect optimization of both properties is difficult to attain from a finite number of data points \citep{wang2020understanding} but can be approximated when the data points (\eg minibatch) are sufficiently large.
Therefore, in addition to \hyperref[eq:densecl]{Eq. (2)}, we also use $\lalign$ and $\luniform$ as the objective functions of the pre-training phase and observe whether the two properties are correlated with the downstream tasks for a wide range of scenarios.
\vspace{-3mm}
\subsection{Dense Feature Matching}
\label{ssec:dfm}
One issue in dense CL is finding the appropriate features to form positive pairs.
The key to matching dense features is that positive pairs must share information (\textit{i.e.} alignment), while negative pairs must repel each other (\textit{i.e.} uniformity). Many studies provide complex strategies to pair strong positives and negative pairs to the anchor \textbf{\emph{$e.g.$}} exploit geometrically identical features \citep{Propa, VADeR}, calculate attention score \citep{SetSim}, or use momentum queue to enlarge the size of negative samples \citep{DenseCL}. We address this issue with a spatially grounded dense feature matching (\textit{i.e.} index-wise matching) based on the assumption from \hyperref[ssec:densecl]{Section 3.2} that dense features of an instance and sampled data points are \textbf{\emph{$i.i.d$}}. Our motivation for doing index-wise matching is to fairly compare the behavior of dense CL on multiple criteria as these tricks could yield various effects for each experiment.

Traditional CL \citep{simclr, swav, misra2020self, moco, selfie, caron2018deep} can learn feature representations when the distance between positive samples is shorter than between negative samples. Also, this approach admits that negative samples contain noisy samples of the positive class, and these noises are negligible when the strong negative samples are large enough. In this context, our simple approach is also reasonable and effective in learning feature representation.
For two dense feature sets $\hb_1 = \{\hb_{(1,1)}, \ldots, \hb_{(1,HW)}\}, \enskip \hb_{(1,i)} \in \mathbb{R}^{d}$ and $\hb_2 = \{\hb_{(2,1)}, \ldots, \hb_{(2,HW)}\}, \enskip \hb_{(2,i)} \in \mathbb{R}^{d}$ from two augmented images, positive pairs are formed by vectors of the same index in each set $pos = \{(\hb_{(1,i)}, \hb_{(2,i)}), \dots, (\hb_{(1,HW)}, \hb_{(2,HW)})\}$ and the vectors of different indices $neg = \tilde\hb_2 \,\text{=}\, \{\hb_{(2,j)}, \ldots, \hb_{(2,HW)}\}, where j \neq i$ are formed as negative pairs including other dense feature vectors from different data points in $\MB$. Therefore, our matching strategy forms a soft positive pair while forming many strong negative pairs ($\approx$ 12.5k dense features of other images; features from different data points) and some noisy negative pairs (different indices from the same data point). Such noisy pairs in negative pairs can be ignored given a large number of strong negative pairs. Although some negative pairs could share information (\textit{e.g.} $\hb_{(1,i)}$ and $\hb_{(2,i+1)}$), asymptotically all negative pairs should follow a uniform distribution.
Surprisingly, this simple matching strategy showed successful performance in all our experiments, suggesting that our $i.i.d$ assumption was not unreasonable.
We further investigate more sophisticated matching strategies that do not make such assumptions: dense feature matching based on cosine similarity \citep{DenseCL}, and set-wise matching based on earth mover distance \citep{SetSim}.
We report in the supplementary that both strategies show either similar or inferior performance to the simple index-wise matching.
\vspace{-2mm}

\section{Experiments}
\vspace{-2mm}
Our experiments primarily focus on the correlation analysis between feature representations after pre-training and the performance of downstream tasks: linear evaluation as the instance-level task and object detection as the dense-level task.
We pose three questions regarding dense features: 1) How does the alignment-uniformity property of dense contrast learning correlate with the performance of object detection and linear evaluation?
2) How different is the behavior of dense feature representations on single or multi-object datasets? 
3) How effective is the index-wise matching strategy in terms of different augmentation techniques?

In this section, we first describe experimental setup and how to quantify the correlation between alignment-uniformity property and downstream task performance.
Then the following three subsections will address each of the three questions above.

\vspace{-3mm}
\subsection{Experimental Setup}
\textbf{Pre-training.}
We conduct pre-training experiments on two datasets: STL-10 \citep{stl10} single-object dataset ($\sim$103k images from the training and unlabeled sets) and MS COCO \citep{coco} multi-object dataset($\sim$118k images from the training set).
We closely follow the hyper-parameters and data augmentation rules from the official implementation of \citet{wang2020understanding} for STL-10 and DenseCL \citep{DenseCL} for COCO.
We use Resnet18 as the backbone and extract the dense features from the penultimate layer (\emph{i.e.} before the global average pooling layer).
Then, these dense features are projected to two different sub-head blocks depending on the training scheme (instance-versus-dense).
We train 200 STL-10 pre-trained models and 120 COCO pre-train models for 200 epochs with instance- and dense-level CL. Each model is optimized with a differently weighted combination of $\lalign$ and $\luniform$, or various values of the temperature $\tau$ of $L_{InfoNCE}$.
Please refer to the supplementary for further details.

\noindent
\textbf{Instance-level Evaluation.}
To evaluate the instance-level linear separation ability, we employ the STL-10 linear evaluation. We freeze the pre-trained weights and fine-tune only one additional linear classification layer for 100 epochs, strictly following the settings of \citet{wang2020understanding}. We use these results as a reference to correlate the instance-level alignment-uniformity properties using the global average pooled feature for each instance.

\noindent
\textbf{Dense-level Evaluation.}
When evaluating dense features, we follow the standard object detection protocol using the Faster R-CNN \citep{ren2015faster} detector (R18-C4 backbone) on the PASCAL VOC trainval 07+12 set and testing on the VOC test 2007 set. Optimization takes a total of 24k iterations. The learning rate is initialized to 0.02 and decayed to be 10 times smaller after 18k and 22k iterations. We use average precision (AP) as an evaluation metric and analyze the correlation by measuring the alignment-uniformity properties of dense features.

\noindent
\textbf{Quantifying Correlation.} 
We quantify the strength of the correlation between alignment-uniformity properties and downstream task performance by utilizing the scalar-valued\\ Kendall's $\tau$, which is a rank-based correlation metric.
Given $\N$ pre-trained models, the two losses ($\lalign$, $\luniform$), and the downstream task performance $P_{task}$ are reordered with min-max normalization across $\N$ models as $r(\lalign)$, $r(\luniform)$, and $r(P_{task})$.
Kendall's $\tau$ correlation metric is
\footnotesize
\begin{equation*}
    \tau \text{=} \frac{P \mathbin{-} Q}{\sqrt{(P + Q + T)(P + Q + U)}}
\end{equation*}
\normalsize
where, $P$ and $Q$ are the numbers of ordered and disordered pairs in $\{r(L_{a_i}) + r(L_{u_i}), r(P_i)\},$ $i \in \N$. $T$ and $U$ are the numbers of ties in $\{r(L_{a_i}) + r(L_{u_i})\}$ and $r(P_i)$, respectively. The correlation value varies between -1 and +1, with a value close to 0 indicating a weak correlation.
Note that a negative correlation between the losses ($\{r(L_{a_i}) + r(L_{u_i})\}$) and downstream task performance ($P_{task}$) indicate that alignment-uniformity are desirable properties, and contrastive pre-training is useful.

\subsection{Results of Pre-training on Single-object Dataset}
\begin{table*}[ht]
\footnotesize
\caption{Single-object dataset results of instance and dense-level evaluation. We show the results for two different training scheme( $L_{InfoNCE}$ and $\lalign \; \& \; \luniform$) in a total of 200 experiments. $\lalign \; \& \; \luniform$ indicates loss of alignment and uniformity.}
\centering
\setlength{\tabcolsep}{3pt}
\vspace{+2mm}
\resizebox{11cm}{!}{\begin{tabular}{lrrrrrrrrrrrrr}
\toprule
    \multirow{3}{*}[-4pt]{Pretraining} &
    \multirow{3}{*}[-4pt]{Loss} &
    \multicolumn{5}{c}{Instance-level Evaluation} & \multicolumn{5}{c}{Dense-level Evaluation} \\
    \cmidrule(lr){3-7}
    \cmidrule(lr){8-12}
    & & \multicolumn{4}{c}{linear evaluation(Acc)} &  \multicolumn{1}{c}{correlation}& \multicolumn{4}{c}{object detection(AP)} & \multicolumn{1}{c}{correlation} \\
    & & exp &   max & Avg&top10 &  $\tau$  &          exp &   max & Avg&top10 &  $\tau$ \\
\midrule

&    $L_{a} \enskip \& \enskip L_{u}$ &    70 & 76.16 & 64.39 & 75.56 &       -0.50 &                      70 & 40.37 & 37.21 & 40.14 & -0.31\\

Instance &    $L_{InfoNCE}$ &          30 & 75.47 & 71.99& 74.97 &       -0.07 &                      30 & 43.38 & 40.17 & 42.33  &       -0.41  \\

 &    $total$            &100 & 76.16 & 66.51&75.61  &       -0.45 &                      100 & 43.38 & 38.02&42.33 &       -0.41 \\
\midrule
& $L_{a} \enskip \& \enskip L_{u}$& 70 & 75.45 & 64.61&75.01  &       -0.19 &                70 & 43.44 & 38.99 & 43.19 &       -0.22\\

Dense   & $L_{InfoNCE}$ & 30 & 75.12 & 60.85& 74.18  &        -0.01  &                        30 & 43.71 & 39.63 & 42.80  &       -0.54 \\

&  $total$ & 100 & 75.45 & 63.47&75.13 &  -0.32 &                    
            100 & 43.71 & 39.2&43.31 &       -0.12  \\
\midrule
Random init            &             &1 & 28.04 & - &        - &                       1 & 31.93 & - &        - \\

\bottomrule
\end{tabular}
\label{tab:singleresult}
}
\end{table*}
\normalsize
\vspace{-3mm}
\begin{figure*}[!ht]
\centering
\includegraphics[width=12.6cm]{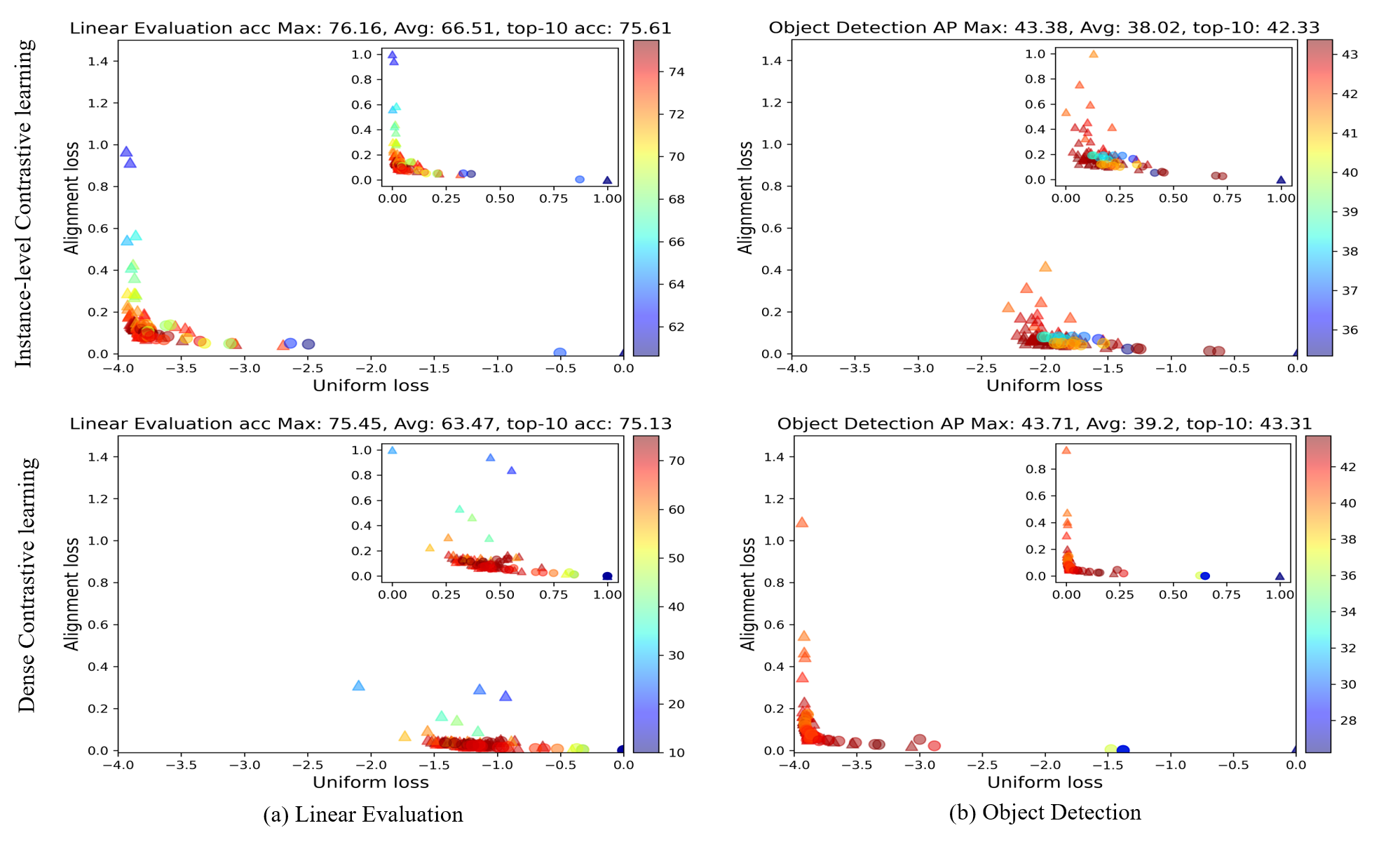}
\footnotesize
\caption{We show the alignment-uniformity property and downstream task performance for each 100 STL10 pre-trained models using instance- or dense-level features. All pre-trained models perform linear evaluation and object detection, then mark each point with color to show the performance. X and Y axes represent uniformity and alignment with a fixed scale. The symbol $\triangle$ and $\circ$ denotes $L_{InfoNCE}$ and $\lalign \; \& \; \luniform$, respectively. We also show normalized $\lalign \; \& \; \luniform$ values in the upper right corner. Note that we examine the alignment-uniformity properties using the features depending on the evaluation aspect (instance vs dense) regardless of the pre-training scheme.} 
\label{fig:exp1}
\end{figure*}
\normalsize
\textbf{Instance-level Evaluation.}
\citet{wang2020understanding} demonstrated that the linear evaluation performance increased with the tendency to optimize alignment-uniformity. Inspired by its findings, we investigate the performance of linear evaluation and alignment-uniformity properties on the STL-10 testset using a global average pooling feature. As shown in \hyperref[fig:exp1]{Fig. 2} (a), the overall trend showed that the linear evaluation performance improved for the optimized alignment-uniformity property in both instance-level and dense CL, and all experiments showed a negative correlation (negative value of $\tau$ in \hyperref[tab:singleresult]{Table 1}). Also, instance-level and dense CL results achieved similar performance with a maximum accuracy of 76.16 and 75.45.
These results show that dense contrast learning pre-trained on a single-object dataset has the ability to linearly separate by capturing the global information. We further investigate the behavior in the object detection task.

\noindent \textbf{Dense-level Evaluation.}
To investigate the dense-level evaluation, we analyze the correlation between the alignment-uniformity of dense features on the STL-10 testset and VOC object detection performance. In this experiment, we can observe that the overall trend of the object detection performance is also correlated with the alignment-uniformity property in both instance-level and dense CL (\hyperref[fig:exp1]{Fig. 2} (b)) . Similar performance was achieved with a maximum AP of 43.38 and 43.71 in both instance-level and dense CL. The instance-level and dense CL using a single object showed a negative correlation between the alignment-uniformity and object detection ability with negative $\tau$ (\hyperref[tab:singleresult]{Table 1}). However, similar trends and performance may have been reached between instance level and dense contrast learning due to the inherent object-centric bias of the STL10 dataset. Still, the gap between the two pre-training schemes remains unknown. Therefore, we perform pre-training on a more complex setup involving multiple objects with the COCO dataset to ensure whether the correlation results of the STL10 pre-training are preserved.

\subsection{Results of Pre-training on Multi-object Dataset.}
\footnotesize
\begin{figure*}[ht!]
\centering
\includegraphics[width=12.6cm]{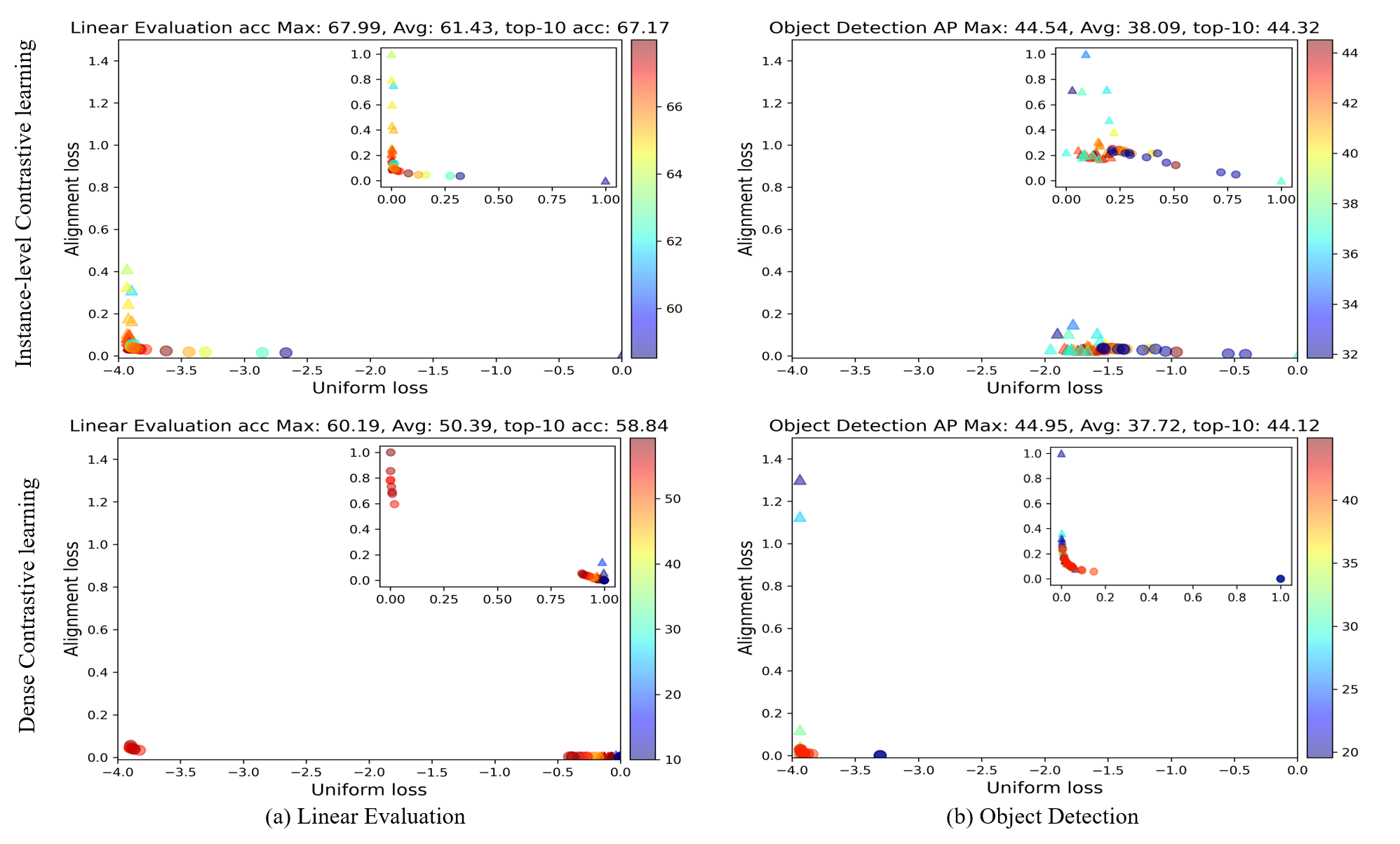}
\caption{We show the alignment-uniformity property and downstream task performance for each 60 COCO pre-trained models using instance- or dense-level features. Each point is marked with color to show its performance and uniformity and alignment properties are represented in X and Y axes with a fixed scale. The symbol $\triangle$ and $\circ$ denotes $L_{InfoNCE}$ and $\lalign \; \& \; \luniform$, respectively.}
\label{fig:exp2}
\end{figure*}
\normalsize
\vspace{-2mm}

\noindent\textbf{Instance-level Evaluation.}
We conduct instance-level evaluations on COCO pre-trained models. The alignment-uniformity properties were measured using the global average pooled feature on the COCO testset while performing linear evaluation using the STL10 dataset. As shown in \hyperref[fig:exp2]{Fig. 3} (a), the trends of instance-level CL showed strong negative correlations with $\tau$ of -0.67 (\hyperref[tab:multiresult]{Table 2}). However, for the pre-training scheme with Dense CL, the results showed an irregular pattern depending on the uniformity, showing a weak correlation of -0.01 tau. Also, COCO pre-training showed inferior to STL pre-training in linear evaluation with maximum accuracy of 67.99 and 60.19 in instance-level and dense CL. We perform an object detection task to investigate whether such a performance gap occurs in dense prediction tasks.

\noindent\textbf{Dense-level Evaluation.}
To evaluate the dense features on COCO pre-trained model, we analyze the correlation between alignment-uniformity of dense features on COCO testset and VOC object detection performance. As seen from \hyperref[fig:exp2]{Fig. 3} (b), all experiments showed high performance as the alignment-uniformity metric decreased. Also, the instance-level and dense CL showed high performance with maximum AP of 44.54 and 44.95 and $\tau$ of -0.21 and -0.13 \hyperref[tab:multiresult]{Table 2}. From these results, pre-training schemes with instance-level or dense-contrast learning using multiple objects perform well in dense prediction tasks despite the complexity of rich semantic information. 
\begin{table}[ht]
\footnotesize
\caption{Multi-object dataset results for instance and dense-level evaluation.}
\centering
\setlength{\tabcolsep}{3pt}
\vspace{+2mm}
\resizebox{11cm}{!}{\begin{tabular}{lrrrrrrrrrrrrrrr}
\toprule
    \multirow{3}{*}[-4pt]{Pretraining} &
    \multirow{3}{*}[-4pt]{Loss} &
    \multicolumn{5}{c}{Instance-level Evaluation} & \multicolumn{5}{c}{Dense-level Evaluation} \\
    \cmidrule(lr){3-7}
    \cmidrule(lr){8-12}
    & & \multicolumn{4}{c}{linear evaluation(Acc)} &  \multicolumn{1}{c}{correlation}& \multicolumn{4}{c}{object detection(AP)} & \multicolumn{1}{c}{correlation} \\
    & & exp &   max & Avg&top10 &  $\tau$  &          exp &   max & Avg&top10 &  $\tau$ \\
\midrule

&    $\lalign \; \& \; \luniform$ &    40 & 67.58 & 59.48 & 66.75 & -0.54 &                      40 & 44.54 & 38.27 & 43.94 &   -0.23\\

Instance &    $L_{InfoNCE}$ &          20 & 67.99 & 65.05 & 66.63 &        -0.67 &                      20 & 44.51 & 37.77 & 42.83 &   -0.03\\

 &    $total$            &60 & 67.99 & 61.43 &67.17 &       -0.67 &                      60 & 44.54 & 38.09&44.32 &      -0.13 \\
\midrule
& $\lalign \; \& \; \luniform$& 40 & 60.19 & 53.41 & 58.64 &  -0.21 &                40 & 44.71 & 36.99& 42.90 &  -0.41\\

Dense   & $L_{InfoNCE}$  &20 & 59.29 & 46.36 & 57.30 &    -0.1 &                        20 & 44.95 & 38.69 & 42.89 &    -0.54 \\
&  $total$ &  60 & 60.19 & 50.39&58.84 & -0.01 &    
            60 & 44.95 & 37.72&44.12 &   -0.21  \\
\midrule
Random init            &             &1 & 28.04 & - &       -& - &                       1 & 31.93 & - &    -&   - \\

\bottomrule
\end{tabular}
\label{tab:multiresult}
}
\end{table}
\vspace{-5mm}
\subsection{Confusing positive samples in Dense CL}
\vspace{-2mm}
\begin{figure*}[ht!]
\centering
\includegraphics[width=12cm]{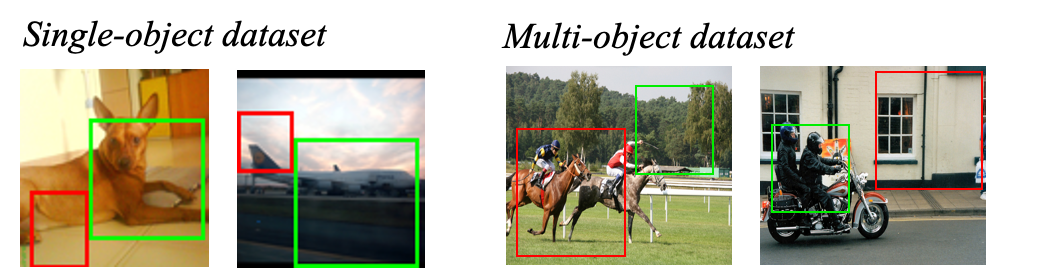}
\caption{Confusing positive samples. The distances between the positive and negative pairs are similar.}
\label{fig:hard}
\end{figure*}
\normalsize
Our assumption of feature matching by the index for positive pairs is that all features are \textbf{\emph{$i.i.d.$}}, but two views from the same image should contain shared information. Single-object datasets, such as STL-10, are discriminated inter-class and object-centered. Due to the innate bias in these data sets, the mutual information in positive pairs (two random views in the same image) naturally shares similar information. However, in more complex setups with multiple objects, such as COCO, there is less chance of sharing semantically identical information even in positive pairs. To further investigate these biases in the data set, we analyze using non-overlapping image settings for confusing positive samples on STL10 and COCO datasets.

\begin{table}[ht]
\footnotesize
\caption{Dense contrastive learning using not-obvious positive samples.}
\centering
\setlength{\tabcolsep}{3pt}
\vspace{+2mm}
\resizebox{11cm}{!}{\begin{tabular}{lrrrrrrrrrrrrrrr}
\toprule
    \multirow{3}{*}[-4pt]{Pretraining} &
    \multirow{3}{*}[-4pt]{} &
    \multicolumn{5}{c}{Instance-level Evaluation} & \multicolumn{5}{c}{Dense-level Evaluation} \\
    \cmidrule(lr){3-7}
    \cmidrule(lr){8-12}
    & & \multicolumn{4}{c}{linear evaluation(Acc)} &  \multicolumn{1}{c}{correlation}& \multicolumn{4}{c}{object detection(AP)} & \multicolumn{1}{c}{correlation} \\
    & & exp &   max & Avg&top10 &  $\tau$  &          exp &   max & Avg&top10 &  $\tau$ \\
\midrule

Single-object &    &          46 & 69.94 & 57.60 & 68.74 & -0.35 &                      46 & 43.06 & 40.01 & 42.78 &   -0.65\\

Multi-object   & & 12 & 54.89 & 40.30 & 44.80 &    -0.15 &                        12 & 32.49 & 32.03 & 32.12 & 0.03 \\

\midrule
Random init            &             &1 & 28.04 & - &       -& - &                       1 & 31.93 & - &    -&   - \\

\bottomrule
\end{tabular}
\label{tab:nonoverlap}
}
\end{table}
In \hyperref[tab:nonoverlap]{Table 3}, the STL10 pre-training results of the linear evaluation and object detection achieved high performance on a single object dataset and showed a strong negative correlation. However, pre-training with confusing positive samples on multi-object datasets showed inferior results in linear evaluation and object detection tasks. In particular, object detection showed similar performance with random initialization result (maximum AP of 31.93) in achieving the maximum AP of 32.49 in the object detection task. It showed a positive correlation with alignment-uniformity ($0.03 \tau$). Therefore, the positive pairing method plays a crucial role in dense contrast learning so that positive pairs can share mutually agreeable information in multi-object datasets. Detailed setup and further experiments are shown in supplementary.
\vspace{-2mm}

\section{Conclusion}
\vspace{-2mm}
In this work, we mainly analyze the theoretical ideas of dense CL using a standard CNN and straightforward feature matching scheme rather than propose a new complex method. By extending existing instance-level CL analysis methods to dense-level, we observe the correlation between alignment-uniformity property of dense features and downstream tasks with newly proposed scalar metrics (linear evaluation and object detection). Also, we discover the core principle in constructing a positive pair of dense features and empirically proved its validity with a simple index-wise matching. In extensive experiments, we find that, regardless of pre-training schemes (instance-level or dense CL), pre-training on single object datasets showed the ability to linearly separate by capturing the global information and perform well on object detection tasks on multiple object datasets. 
Furthermore, our work can be potentially used to compare the performance of different CL schemes by evaluating alignment-uniformity properties of instance- and dense-level features before performing downstream tasks.
The novelty of our work lies in carefully designed experiments and evaluation metric, allowing a reliable conversion from the ``expected'' to ``confirmed''. We believe that the researchers can now safely rely on our findings and move on to developing more principled CL methods in the future, while treating our methods as a minimum baseline.
\newpage

\section*{Acknowledgements}
This work was supported by the KAIST-NAVER Hyper-Creative AI Center and the Institute of Information \& Communications Technology Planning \& Evaluation (IITP) grants (No.2019-0-00075 Artificial Intelligence Graduate School Program(KAIST) and No.2022-0-009840101003), and National Research Foundation of Korea (NRF) grant\\
(NRF-2020H1D3A2A03100945) funded by the Korea government (MSIT).

\bibliography{egbib}
\newpage
\section{Supplementary}
\subsection{Experimental setup details}
\noindent
\textbf{Architecture.} We use Resnet18 as the backbone and extract the dense features from the penultimate layer. Then, these dense features (\emph{512-dim}) are projected to two different sub-head blocks depending on the training scheme (instance-versus-dense). For the instance feature embedding, the projection head consists of a global pooling layer and the configuration of \emph{MLP(512-dim)-ReLU-MLP(128-dim)}. However, dense feature embedding removes the global pooling layer and replaces the MLP head with a 1 x 1 convolution layer to keep the spatial information: the configuration of \emph{Conv(512-dim)-ReLU-Conv(128-dim)}. 

\vspace{+5mm}
\noindent
\textbf{Pretraining setup.}
We conduct pretraining experiments following the data augmentation rule of \citet{wang2020understanding} for STL10 pretraining: random horizontal flip, random color jittering, random grayscale conversion, and $64 \times 64$ pixel crop with the scale 0.08 to 1.0 of the original image (an average 0.6 intersection ratio between two cropped images.) We use SGD as our optimizer with the learning rate decayed by a factor of 0.1 at epochs 155, 170, and 185. The SGD momentum is set to 0.9. For COCO pretraining, we follow \citet{DenseCL} to adopt data augmentation with random horizontal flip, random color jittering, random grayscale conversion, and $224 \times 224$ pixel crop is taken with the scale 0.2 to 1.0  of the original image (an average 0.7 intersection ratio between two cropped image). We adopt SGD as the optimizer and set its weight decay and momentum to 0.0001 and 0.9 with a reciprocal learning rate decay schedule (warm-up iteration set to 0). All experiments are performed on 4-8 2080 Ti GPUs, RTX-3090 GPUs, and RTX-A6000 GPUs.

\vspace{+5mm}
\noindent
\textbf{Evaluation protocol.}
We adopt augmentation rule to measure the alignment-uniformity property in stl10 and coco datasets: 
\begin{itemize}
    \item alignment: Random resized crop with the scale 0.95 to 1.0 of the original image, color jittering, and random grayscale conversion.
    \item uniformity: Resize and centercrop.
\end{itemize}

\noindent
We measure the $L_{a}$ and $L_{u}$ properties for instance-level and density features. All features were $L2$-normalized, as the metrics are defined on the hypersphere. For instance-level evaluation, each instance is transformed to a global averaged pooled feature by $\fb$ and then measured the alignment and uniformity properties in $\MB$. For dense-level evaluation, each instance is transformed to a dense feature set by $\fb$ and then measures the alignment and uniformity properties in $\MB$. For linear evaluation details, we follow the standard linear evaluation on the STL10 protocol and report results on the validation set. We report performance after learning linear classifiers for 100 epochs, with an initial learning rate of 0.001, a batch size of 128, and a step learning rate schedule that drops at epochs 60 and 80 with the Adam optimizer.

\subsection{Dense Feature Matching.}
In our experiment, we assumes all dense features are $i.i.d.$. Namely, we consider a one-to-one relationship. To explore the one-to-many and many-to-many relationships, we investigate sophisticated matching strategies with $infoNCE$ loss ($L_c$): 1) one-to-many: dense feature matching based on cosine similarity \cite{DenseCL}, and 2) many-to-many: set-wise matching based on earth mover distance. 

\subsubsection{One-to-Many Feature Matching}
Inspired by the \citet{DenseCL}, we perform maximum cosine similarity feature matching. As all our experimental settings are similar to SimCLR, we extract feature maps from a single encoder $\fb$ and compute cosine similarity matching in $\MB$. This setting considers a one-to-many relationship. Specifically, after two augmented views $x$ and $x'$ are fed to $\fb$ from the same input image, $\hb_1=f(x)=\{\hb_1, \hb_2, \ldots, \hb_{HW}\}, \enskip \hb_i \in \mathbb{R}^{d}$, and $\hb_2=f(x')=\{\hb_1, \hb_2, \ldots, \hb_{HW}\}, \enskip \hb_i \in \mathbb{R}^{d}$ are acquired, where ${HW}$ is the spatial dimension size. Then, each dense feature of $\hb_1$ retrieves the maximum cosine similarity value in $\hb_2$ as a positive pair. Therefore, the number of positive pairs in each instance equals the number of dense features ($\hb_1$). For negative pairs, $\hb_1$ computes cosine similarity with the dense features of other instances in $\MB$, pushing each other.

\begin{align*}
    positive = argmax_{(i,j)} sim(\hb_{(i,j)},\hb_{(i,j)}),\\
    neagtive = argmax_{(i,j)}\sum_{k\neq i}^\N sim(\hb_{(i,j)},\hb_{(k,j)}),
\end{align*}

We train the models on the STL10 and COCO datasets and evaluate linear evaluation and object detection tasks. As suggested by DenseCL, we also pre-train the model with a weighting ratio of 0.5 for instance-level (global mean pooling) and dense features. We emphasize that both linear evaluation and object detection fail (\hyperref[tab:densecl]{Table 4} Cos with $L_d$) when using dense features as the only training vector for computing cosine similarity in the SimCLR setup. This shows that mode collapse occurs during pre-training and proves that cosine matching is a suboptimal method for dense feature mapping.

\subsubsection{Many-to-Many Feature Matching}
We show a many-to-many matching method of dense features. Set-wise dense feature matching is recently studied \citet{SetSim} because the dense-level correspondence tends to be noisy because of many similar misleading features, \eg backgrounds. We focus on this set-wise matching method and leverage earth mover distance (\emph{$i.e.$} optimal transport problem) for dense feature matching. Earth Mover's Distance (EMD) \cite{rubner2000earth, otter} is a many-to-many matching method in which the individual element distances are constructed as the distances between two sets of distributions, the discrete form of which can be formulated as an optimal transport problem. Specially, for two feature maps $\hb_1$, $\hb_2$, EMD between two feature maps as the minimum \emph{transport cost} from $\hb_1$ to $\hb_2$. 

\begin{align*}
    U(r,c) \textbf{=} \{P \in \mathbb{R}^{HW\times HW}| P\mathbbm{1} \textbf{=} \mathbf{r}, P^T\mathbbm{1} \textbf{=} \mathbf{c}\}.
\end{align*}

where, $\mathbbm{1} \in \mathbb{R}^{HW}$ are the vectors of all ones. $\mathbf{r}$ and $\mathbf{c}$ are marginal weights of matrix $P$ onto its rows and columns, respectively. Then, for the transport cost map ($\mathcal{TM}$), we utilized the cosine distance between $\hb_1$ and $\hb_2$. EMD is defined as follows:

\begin{align*}
    EMD(r,c) \enskip \textbf{=}\enskip \min_{P \in U(r,c)} <P, \mathcal{TM}>
\end{align*}

where, $\mathcal{TM}$ is the cosine distance matrix between $\hb_1$ and $\hb_2$ and <$\cdot$, $\cdot$> stands for the Frobenius dot-product between two matrices.

We calculate the optimal transport using a fast iterative solution named \emph{Sinkhorn-Knopp algorithm} with a regularization term $E = 0.1$ as:
\begin{align*}
    \min_{P \in U(r,c)} <P, \mathcal{TM}> + \frac{1}{\lambda}E(P),
\end{align*}

where $E(P) \textit{=} P(logP \mathrm{-} 1)$ and $\lambda$ is a constant hyper-parameter that controls the intensity of regularization term. The approximated optimal transprot plan $P^* = diag(v)\times P \times diag(u)$, where $P = e^{-\lambda \mathcal{TM}}$ is the element-wise exponential of $-\lambda \mathcal{TM}$ and $v$ and $u$ are two vectors of scaling coefficients chosen so that the resulting matrix $P \in U(r,c)$. The vector $u$ and $v$ can be obtanied via a simple iteration as follows:
\begin{align*}
    \forall i, \enskip v_i^{n+1} &\leftarrow \frac{r_i}{\sum_jP_{i,j}u_j^n}\\
    \forall j, \enskip u_j^{n+1} &\leftarrow \frac{c_j}{\sum_iP_{i,j}v_j^{n+1}}
\end{align*}

After iterate $N = 10$ times, $P^*$ can be obtained. Finally, we can compute the similarity score $OT_{distance}$ between two dense features ($\hb_1$ and $\hb_2$) with:

\begin{align*}
    OT_{distance} = <P, \mathcal{TM}>
\end{align*}

Despite this complex matching process and computational overhead, we find that the STL10 and COCO pretraining obtained inferior results in the linear evaluation and comparable to our index-wise matching in object detection (\hyperref[tab:densecl]{Table 4}). Therefore, we believe that index-wise matching method is straightforward and reasonable without additional computational overhead.

\begin{table}[ht]
\caption{Dense feature matching. $L_{i}$ and $L_{d}$ indicates instance-level and dense contrastive learning. $L_{i}$ + $L_{d}$ represent pre-training with weight ratios of 0.5 each. Cos and OT denote matching methods with cosine similarity and optimal transport. Cos (COCO) and OT (COCO) experiments used same  hyperparameters with \citet{DenseCL}. We show our index-wise matching (Ind) results.} \label{tab:densecl}
\vspace{+1mm}
\centering
\setlength{\tabcolsep}{3pt}
\resizebox{11cm}{!}{\begin{tabular}{lcrrrrrrrr}
\toprule
{} &{} & \multicolumn{4}{c}{linear evaluation (Acc)} & \multicolumn{4}{c}{object detection (AP)} \\
{}  & Loss &exp &   max &  mean & top10 & exp &   max &  mean & top10 \\
\midrule
Cos (STL10) & $L_{i}$ + $L_d$ &20 & 11.92 & 10.2 & 10.4 & 20 & 43.68 & 41.34 & 42.68\\
Cos (STL10) & $L_d$ & 20 & 10 & 10 & 10 & 20 & 28.49 & 1.42 & 2.85\\
OT (STL10)& $L_d$  & 20 & 59.32 & 30.66 & 38.9 & 20 & 40.4 & 39.44 & 39.81\\
Cos (COCO) & $L_{i}$ + $L_d$ & 1 & 22.82 & 22.82 & - & 1 & 43.83 & 43.83 & -\\
OT (COCO) & $L_d$ & 1 & 22.85 &  22.85 & - & 1 & 43.39 & 43.39 & -\\
\midrule
Ind (stl10) & $L_d$ & 100 & 75.45 & 63.47&75.13 & 100 & 43.71 & 39.2&43.31\\
Ind (coco)  & $L_d$  &  60 & 60.19 & 50.39&58.84 & 60 & 44.95 & 37.72&44.12\\
\bottomrule
\end{tabular}
}
\end{table}
\normalsize

\subsection{Detailed Results.}

We show the detailed pretraining phase and downstream task results. Specifically, we show the hyperparameter settings for batch size, learning rate, ratio of loss weights between $L_a, L_u,$ and $L_c$ during pretraining, and normalized temperature $\tau$ for $L_c$. In addition, each pre-trained model shows the results of Instance-level versus Dense-level evaluation (metrics for $L_a, L_u,$ and downstream task performance) according to two evaluation aspects. \hyperref[tab:supplestlinst]{Table 5} shows 100 STL pretraining based on instace-level contrastive learning, \hyperref[tab:supplestldense]{Table 6} shows 100 STL pretraining based on dense contrastive learning. Also, \hyperref[tab:supplecocoinst]{Table 7} and \hyperref[tab:supplecocodense]{Table 8} show 60 coco pretraining based on instace-level contrastive learning and 60 coco pretraining based on dense contrastive learning. 
Next, we show the results of confusing positive pairing in a non-overlapping setting on STL10 (\hyperref[tab:hardstl10]{Table 9}) and COCO (\hyperref[tab:hardcoco]{Table 10}) dataset.

\scriptsize
\begin{longtable}{lcccccccccccc}
    \caption{100 STL10 pretraining: Instance-level contrastive learning.} \label{tab:supplestlinst} \\
    \toprule
    \multicolumn{6}{c}{Pretraining} & \multicolumn{3}{c}{Instance-level Evaluation} & \multicolumn{3}{c}{Dense-level Evaluation} \\
    \cmidrule(lr){1-6}
    \cmidrule(lr){7-9}
    \cmidrule(lr){10-12}
    Batch &    LR & $L_a$ & $L_u$ & $L_{c}$ & $L_{c}/\tau$  & $L_a$ & $L_u$ &  acc &$L_a$ & $L_u$ &ap \\
    \midrule
    \endfirsthead

    \multicolumn{7}{c}{{\bfseries \tablename\ \thetable{} -- continued from previous page}} \\
    \toprule
    \multicolumn{6}{c}{Pretraining} & \multicolumn{3}{c}{Instance-level Evaluation} & \multicolumn{3}{c}{Dense-level Evaluation} \\
    \cmidrule(lr){1-6}
    \cmidrule(lr){7-9}
    \cmidrule(lr){10-12}
    Batch &    LR & $L_a$ & $L_u$ & $L_{c}$ & $L_{c}/\tau$  & $L_a$ & $L_u$ &  acc &$L_a$ & $L_u$ &ap \\
    \midrule
    \endhead

    \midrule
    \multicolumn{8}{r}{{Continued on next page}} \\ 
    \midrule
    \endfoot

    \bottomrule
    \endlastfoot
768 & 0.36 &  0.3000 & 0.400 &           1.0 &       0.07 &  0.1345 & -3.5469 & 68.0250 &                 0.0825 & -1.7645 & 36.5862 \\
768 & 0.36 &  0.3000 & 0.400 &           1.0 &       0.10 &  0.1346 & -3.7992 & 73.1250 &                 0.0874 & -1.9413 & 38.0914 \\
768 & 0.36 &  0.0200 & 0.980 &           0.0 &       0.00 &  0.2337 & -3.9243 & 59.2000 &                 0.0936 & -1.8838 & 37.4927 \\
768 & 0.36 &  0.0000 & 1.000 &           0.0 &       0.00 &  0.9691 & -3.9344 & 16.6000 &                 0.4179 & -1.9928 & 31.0784 \\
768 & 0.36 &  0.1000 & 1.800 &           0.0 &       0.00 &  0.1819 & -3.9127 & 66.4375 &                 0.0662 & -1.7201 & 38.7480 \\
768 & 0.36 &  0.5000 & 0.000 &           1.0 &       0.50 &  0.0489 & -3.0705 & 72.2875 &                 0.0340 & -1.5140 & 39.2179 \\
768 & 0.36 &  0.4000 & 0.200 &           1.0 &       0.10 &  0.1230 & -3.7343 & 72.3625 &                 0.0780 & -1.8185 & 37.4842 \\
768 & 0.36 &  0.0750 & 1.850 &           0.0 &       0.00 &  0.1947 & -3.8887 & 63.6250 &                 0.0831 & -1.8285 & 38.9777 \\
\end{longtable}
\normalsize 

\scriptsize
\begin{longtable}{lcccccccccccc}
    \caption{100 STL10 pretraining: Dense contrastive learning.} \label{tab:supplestldense} \\
    \toprule
    \multicolumn{6}{c}{Pretraining} & \multicolumn{3}{c}{Instance-level Evaluation} & \multicolumn{3}{c}{Dense-level Evaluation} \\
    \cmidrule(lr){1-6}
    \cmidrule(lr){7-9}
    \cmidrule(lr){10-12}
    Batch &    LR & $L_a$ & $L_u$ & $L_{c}$ & $L_{c}/\tau$  & $L_a$ & $L_u$ &  acc &$L_a$ & $L_u$ &ap \\
    \midrule
    \endfirsthead

    \multicolumn{7}{c}{{\bfseries \tablename\ \thetable{} -- continued from previous page}} \\
    \toprule
    \multicolumn{6}{c}{Pretraining} & \multicolumn{3}{c}{Instance-level Evaluation} & \multicolumn{3}{c}{Dense-level Evaluation} \\
    \cmidrule(lr){1-6}
    \cmidrule(lr){7-9}
    \cmidrule(lr){10-12}
    Batch &    LR & $L_a$ & $L_u$ & $L_{c}$ & $L_{c}/\tau$  & $L_a$ & $L_u$ &  acc &$L_a$ & $L_u$ &ap \\
    \midrule
    \endhead

    \midrule
    \multicolumn{8}{r}{{Continued on next page}} \\ 
    \midrule
    \endfoot

    \bottomrule
    \endlastfoot
768 &  0.36 &  0.4000 &  0.200 &           1.0 &       0.50 &  0.0116 & -0.8361 &  70.6500 &                 0.0355 & -3.5337 &  43.2070 \\
768 &  0.36 &  0.0125 &  1.000 &           0.0 &       0.00 &  0.0395 & -1.1361 &  58.1250 &                 0.1884 & -3.9308 &  39.7316 \\
768 &  0.36 &  0.1000 &  0.800 &           1.0 &       0.10 &  0.0402 & -1.3417 &  74.5875 &                 0.1196 & -3.9134 &  39.9886 \\
768 &  0.36 &  0.4000 &  1.200 &           0.0 &       0.00 &  0.0350 & -1.2664 &  73.7750 &                 0.0949 & -3.8995 &  43.1266 \\
768 &  0.36 &  0.3000 &  0.400 &           1.0 &       0.07 &  0.0413 & -1.0062 &  72.8875 &                 0.1596 & -3.8974 &  39.7622 \\
768 &  0.36 &  0.4000 &  1.000 &           0.0 &       0.00 &  0.0363 & -1.3714 &  74.7250 &                 0.0926 & -3.8955 &  43.4382 \\
768 &  0.36 &  0.3000 &  0.400 &           1.0 &       0.50 &  0.0145 & -0.9059 &  72.2500 &                 0.0451 & -3.6832 &  42.9837 \\
768 &  0.36 &  0.2500 &  1.000 &           0.0 &       0.00 &  0.0349 & -1.2652 &  73.6625 &                 0.1033 & -3.9044 &  43.0176 \\
\end{longtable}
\normalsize 

\scriptsize
\begin{longtable}{lcccccccccccc}
    \caption{60 COCO pretraining: Instance-level contrastive learning.} \label{tab:supplecocoinst} \\
    \toprule
    \multicolumn{6}{c}{Pretraining} & \multicolumn{3}{c}{Instance-level Evaluation} & \multicolumn{3}{c}{Dense-level Evaluation} \\
    \cmidrule(lr){1-6}
    \cmidrule(lr){7-9}
    \cmidrule(lr){10-12}
    Batch &    LR & $L_a$ & $L_u$ & $L_{c}$ & $L_{c}/\tau$  & $L_a$ & $L_u$ &  acc &$L_a$ & $L_u$ &ap \\
    \midrule
    \endfirsthead

    \multicolumn{7}{c}{{\bfseries \tablename\ \thetable{} -- continued from previous page}} \\
    \toprule
    \multicolumn{6}{c}{Pretraining} & \multicolumn{3}{c}{Instance-level Evaluation} & \multicolumn{3}{c}{Dense-level Evaluation} \\
    \cmidrule(lr){1-6}
    \cmidrule(lr){7-9}
    \cmidrule(lr){10-12}
    Batch &    LR & $L_a$ & $L_u$ & $L_{c}$ & $L_{c}/\tau$  & $L_a$ & $L_u$ &  acc &$L_a$ & $L_u$ &ap \\
    \midrule
    \endhead

    \midrule
    \multicolumn{8}{r}{{Continued on next page}} \\ 
    \midrule
    \endfoot

    \bottomrule
    \endlastfoot
128 &  0.15 &  1.0250 &  1.000 &           0.0 &        0.0 &  0.0387 & -3.8838 &  67.1500 &                 0.0311 & -1.7023 &  43.9544 \\
128 &  0.15 &  0.3000 &  1.000 &           0.0 &        0.0 &  0.0444 & -3.9045 &  66.5125 &                 0.0272 & -1.7377 &  44.1403 \\
128 &  0.15 &  0.0500 &  1.000 &           0.0 &        0.0 &  0.0665 & -3.9135 &  59.7625 &                 0.0314 & -1.7136 &  42.0488 \\
128 &  0.15 &  1.0000 &  0.975 &           0.0 &        0.0 &  0.0376 & -3.8859 &  67.5625 &                 0.0304 & -1.6904 &  44.3160 \\
128 &  0.15 &  0.2500 &  1.500 &           0.0 &        0.0 &  0.0505 & -3.9071 &  64.4750 &                 0.0309 & -1.8215 &  43.9035 \\
128 &  0.15 &  1.0000 &  1.000 &           0.0 &        0.0 &  0.0373 & -3.8842 &  65.9250 &                 0.0308 & -1.7046 &  44.5365 \\
128 &  0.15 &  0.0000 &  5.000 &           0.0 &        0.0 &  0.3122 & -3.8956 &  29.7500 &                 0.1078 & -1.8993 &  24.0097 \\
128 &  0.15 &  0.0750 &  1.850 &           0.0 &        0.0 &  0.0710 & -3.9145 &  60.0875 &                 0.0363 & -1.8449 &  41.5761 \\
\end{longtable}
\normalsize 
\scriptsize
\begin{longtable}{lcccccccccccc}
    \caption{60 COCO pretraining: Dense contrastive learning.} \label{tab:supplecocodense} \\
    \toprule
    \multicolumn{6}{c}{Pretraining} & \multicolumn{3}{c}{Instance-level Evaluation} & \multicolumn{3}{c}{Dense-level Evaluation} \\
    \cmidrule(lr){1-6}
    \cmidrule(lr){7-9}
    \cmidrule(lr){10-12}
    Batch &    LR & $L_a$ & $L_u$ & $L_{c}$ & $L_{c}/\tau$  & $L_a$ & $L_u$ &  acc &$L_a$ & $L_u$ &ap \\
    \midrule
    \endfirsthead

    \multicolumn{7}{c}{{\bfseries \tablename\ \thetable{} -- continued from previous page}} \\
    \toprule
    \multicolumn{6}{c}{Pretraining} & \multicolumn{3}{c}{Instance-level Evaluation} & \multicolumn{3}{c}{Dense-level Evaluation} \\
    \cmidrule(lr){1-6}
    \cmidrule(lr){7-9}
    \cmidrule(lr){10-12}
    Batch &    LR & $L_a$ & $L_u$ & $L_{c}$ & $L_{c}/\tau$  & $L_a$ & $L_u$ &  acc &$L_a$ & $L_u$ &ap \\
    \midrule
    \endhead

    \midrule
    \multicolumn{8}{r}{{Continued on next page}} \\ 
    \midrule
    \endfoot

    \bottomrule
    \endlastfoot
    128 &  0.15 &  1.0000 &  0.975 &           0.0 &        0.0 &  0.0010 & -0.1481 &  50.6625 &                 0.0118 & -3.9018 &  44.4962 \\
    128 &  0.15 &  1.0250 &  1.000 &           0.0 &        0.0 &  0.0010 & -0.1550 &  51.4875 &                 0.0114 & -3.9009 &  44.7054 \\
    128 &  0.15 &  0.3000 &  1.000 &           0.0 &        0.0 &  0.0005 & -0.0706 &  57.1750 &                 0.0179 & -3.9244 &  44.1765 \\
    128 &  0.15 &  0.0500 &  1.000 &           0.0 &        0.0 &  0.0006 & -0.0804 &  58.1500 &                 0.0279 & -3.9332 &  39.5766 \\
    128 &  0.15 &  0.2500 &  1.500 &           0.0 &        0.0 &  0.0006 & -0.0710 &  58.9500 &                 0.0210 & -3.9297 &  43.5093 \\
    128 &  0.15 &  0.0750 &  1.850 &           0.0 &        0.0 &  0.0006 & -0.0762 &  58.7375 &                 0.0300 & -3.9343 &  41.3990 \\
    128 &  0.15 &  0.0000 &  5.000 &           0.0 &        0.0 &  0.0034 & -0.0089 &  15.3875 &                 1.3034 & -3.9380 &  20.8865 \\
    128 &  0.15 &  1.0000 &  1.000 &           0.0 &        0.0 &  0.0009 & -0.1446 &  50.2125 &                 0.0113 & -3.8989 &  44.6067 \\
\end{longtable}
\normalsize 
\scriptsize
\begin{longtable}{lcccccccccccc}
    \caption{Confusing positive paring on Single-object Dataset.} \label{tab:hardstl10} \\
    \toprule
    \multicolumn{6}{c}{Pretraining} & \multicolumn{3}{c}{Instance-level Evaluation} & \multicolumn{3}{c}{Dense-level Evaluation} \\
    \cmidrule(lr){1-6}
    \cmidrule(lr){7-9}
    \cmidrule(lr){10-12}
    Batch &    LR & $L_a$ & $L_u$ & $L_{c}$ & $L_{c}/\tau$  & $L_a$ & $L_u$ &  acc &$L_a$ & $L_u$ &ap \\
    \midrule
    \endfirsthead

    \multicolumn{7}{c}{{\bfseries \tablename\ \thetable{} -- continued from previous page}} \\
    \toprule
    \multicolumn{6}{c}{Pretraining} & \multicolumn{3}{c}{Instance-level Evaluation} & \multicolumn{3}{c}{Dense-level Evaluation} \\
    \cmidrule(lr){1-6}
    \cmidrule(lr){7-9}
    \cmidrule(lr){10-12}
    Batch &    LR & $L_a$ & $L_u$ & $L_{c}$ & $L_{c}/\tau$  & $L_a$ & $L_u$ &  acc &$L_a$ & $L_u$ &ap \\
    \midrule
    \endhead

    \midrule
    \multicolumn{8}{r}{{Continued on next page}} \\ 
    \midrule
    \endfoot

    \bottomrule
    \endlastfoot
768 &  0.36 &  0.2000 &  1.600 &           0.0 &       0.00 &  0.2146 & -1.4416 &  65.8125 &                 0.5255 & -3.9088 &  41.8164 \\
768 &  0.36 &  0.3000 &  0.400 &           1.0 &       0.07 &  0.1306 & -0.9935 &  66.5750 &                 0.3819 & -3.7868 &  38.4735 \\
768 &  0.36 &  1.0000 &  0.000 &           1.0 &       0.50 &  0.0000 &  0.0000 &  10.4125 &                 0.0000 & -1.3778 &  32.1306 \\
768 &  0.36 &  0.0000 &  5.000 &           0.0 &       0.00 &  0.2985 & -0.7778 &  14.0625 &                 1.5406 & -3.9183 &  30.7468 \\
768 &  0.36 &  1.0000 &  2.000 &           0.0 &       0.00 &  0.1624 & -1.3117 &  67.5750 &                 0.4095 & -3.8566 &  42.1929 \\
768 &  0.36 &  0.2000 &  0.600 &           1.0 &       0.50 &  0.0705 & -0.9362 &  65.2000 &                 0.2099 & -3.6027 &  42.1262 \\
768 &  0.36 &  0.4000 &  0.200 &           1.0 &       0.07 &  0.1095 & -0.9037 &  65.1750 &                 0.3550 & -3.7344 &  38.4746 \\
768 &  0.36 &  0.3000 &  0.400 &           1.0 &       0.10 &  0.1201 & -1.0540 &  69.1250 &                 0.3904 & -3.8408 &  39.4105 \\
\end{longtable}
\normalsize 
\scriptsize
\begin{longtable}{lcccccccccccc}
    \caption{Confusing positive paring on Multi-object Dataset.} \label{tab:hardcoco} \\
    \toprule
    \multicolumn{6}{c}{Pretraining} & \multicolumn{3}{c}{Instance-level Evaluation} & \multicolumn{3}{c}{Dense-level Evaluation} \\
    \cmidrule(lr){1-6}
    \cmidrule(lr){7-9}
    \cmidrule(lr){10-12}
    Batch &    LR & $L_a$ & $L_u$ & $L_{c}$ & $L_{c}/\tau$  & $L_a$ & $L_u$ &  acc &$L_a$ & $L_u$ &ap \\
    \midrule
    \endfirsthead

    \multicolumn{7}{c}{{\bfseries \tablename\ \thetable{} -- continued from previous page}} \\
    \toprule
    \multicolumn{6}{c}{Pretraining} & \multicolumn{3}{c}{Instance-level Evaluation} & \multicolumn{3}{c}{Dense-level Evaluation} \\
    \cmidrule(lr){1-6}
    \cmidrule(lr){7-9}
    \cmidrule(lr){10-12}
    Batch &    LR & $L_a$ & $L_u$ & $L_{c}$ & $L_{c}/\tau$  & $L_a$ & $L_u$ &  acc &$L_a$ & $L_u$ &ap \\
    \midrule
    \endhead

    \midrule
    \multicolumn{8}{r}{{Continued on next page}} \\ 
    \midrule
    \endfoot

    \bottomrule
    \endlastfoot
128 &  0.15 &     0.0 &   0.0 &           1.0 &      0.300 &  0.1383 & -3.5340 &  36.1875 &                 0.0101 & -3.6290 &  32.3559 \\
128 &  0.15 &     0.0 &   0.0 &           1.0 &      1.000 &  0.0640 & -2.7377 &  18.5875 &                 0.0000 & -3.3034 &  32.4726 \\
128 &  0.15 &     0.0 &   0.0 &           1.0 &      0.130 &  0.2090 & -3.7431 &  54.5000 &                 0.0779 & -3.9079 &  32.0033 \\
128 &  0.15 &     0.0 &   0.0 &           1.0 &      0.190 &  0.1730 & -3.6946 &  48.0750 &                 0.0285 & -3.8384 &  31.6413 \\
128 &  0.15 &     0.0 &   0.0 &           1.0 &      0.500 &  0.0999 & -3.2697 &  19.3500 &                 0.0000 & -3.3034 &  31.9132 \\
128 &  0.15 &     0.0 &   0.0 &           1.0 &      0.100 &  0.2132 & -3.7242 &  41.0250 &                 0.1203 & -3.9243 &  32.4916 \\
128 &  0.15 &     0.0 &   0.0 &           1.0 &      0.150 &  0.1926 & -3.7301 &  53.9125 &                 0.0524 & -3.8928 &  31.8794 \\
128 &  0.15 &     0.0 &   0.0 &           1.0 &      0.750 &  0.0745 & -2.9549 &  17.0250 &                 0.0000 & -3.3034 &  31.6898 \\
\end{longtable}
\normalsize 

\end{document}